\documentclass[11pt]{article}

\usepackage[utf8]{inputenc}
\usepackage[T1]{fontenc}
\usepackage[margin=1in]{geometry}
\usepackage{graphicx}
\usepackage{booktabs}
\usepackage{multirow}
\usepackage{amsmath}
\usepackage{amssymb}
\usepackage{textcomp}
\usepackage{listings}
\usepackage{xcolor}
\usepackage{float}
\usepackage{caption}
\usepackage{url}
\usepackage[numbers,sort&compress]{natbib}
\usepackage[hidelinks]{hyperref}

\DeclareUnicodeCharacter{2248}{\ensuremath{\approx}}
\DeclareUnicodeCharacter{2014}{---}
\DeclareUnicodeCharacter{2013}{--}
\DeclareUnicodeCharacter{2019}{'}
\DeclareUnicodeCharacter{201C}{``}
\DeclareUnicodeCharacter{201D}{''}
\DeclareUnicodeCharacter{2192}{$\rightarrow$}

\lstdefinestyle{promptstyle}{
    basicstyle=\ttfamily\scriptsize,
    breaklines=true,
    breakatwhitespace=false,
    columns=fullflexible,
    frame=single,
    backgroundcolor=\color{gray!5},
    xleftmargin=2pt,
    xrightmargin=2pt,
    showstringspaces=false
}

\title{\textbf{Industrial-Instruction: An End-to-End Framework for Building
Instruction-Tuning and Benchmark Datasets from Industrial Technical Reports}}

\author{
  Parsa Bakhtiari\thanks{Department of Computer Engineering, Hamedan University of Technology, Hamedan, Iran. \texttt{p.bakhtiari@stu.hut.ac.ir}} \and
  Hassan Bashiri\thanks{Department of Computer Engineering, Hamedan University
  of Technology, Hamedan, Iran. \texttt{bashiri@hut.ac.ir}} \and
  Alireza Khalilipour\thanks{Department of Computer Science, University of
  Antwerp, Antwerp, Belgium; Flanders Make Strategic Research Center, Antwerp,
  Belgium. \texttt{alireza.khalilipour@uantwerpen.be}} \and
  Masoud Nasiripour\thanks{Department of Computer Engineering, Hamedan
  University of Technology, Hamedan, Iran. \texttt{masoud.nasiri@stu.hut.ac.ir}} \and
  Moharram Challenger\thanks{Department of Computer Science, University of
  Antwerp, Antwerp, Belgium; Flanders Make Strategic Research Center, Antwerp,
  Belgium. \texttt{moharram.challenger@uantwerpen.be}}
}
\date{}

\begin{document}
\maketitle

\begin{abstract}
Industrial technical reports contain high-value knowledge for maintenance, troubleshooting, and product engineering, but their heterogeneous structure (dense prose, specifications, and tables) makes them difficult to index and reason over with standard retrieval and question-answering pipelines, and no public instruction-tuning or benchmark datasets are built from such documents. We address this gap with Industrial-Instruction, contributing (i) two open question-answering datasets built from real industrial technical reports and (ii) the end-to-end pipeline that produces them. Using 906 publicly available Panasonic documents (7,525 pages) as a case study, we apply layout-aware extraction to preserve text and tabular content, construct a semantic index for retrieval, and synthesize multiple-choice QA instances grounded in retrieved evidence. Each dataset explicitly models five realistic query--document relationships---irrelevant retrieval, single-/multi-document support, and single-/multi-document answer---so that models can be trained and evaluated on robustness to retrieval noise and on multi-step evidence integration. After automated filtering of an initial 23.9k generated samples, each dataset provides $\approx$13.6k QA pairs with associated source documents and a held-out benchmark split. Experiments with small open LLMs ($<$10B parameters) show that full fine-tuning on Industrial-Instruction substantially improves performance on the Panasonic benchmark, increasing Set-Match Accuracy from 28.5\% to 42.0\% and F1 from 46.6\% to 63.5\%, with consistent gains both with and without RAG.

We release the dataset in two parallel versions built by the same pipeline---one generated with the open-weight Qwen3-30B-A3B-Instruct model and one with the closed, API-based Claude-Opus-4.6 model---which additionally enables a direct comparison of open- versus frontier-model data generation. The Claude-Opus-4.6-generated dataset produces a substantially cleaner raw corpus and larger downstream fine-tuning gains, at roughly two orders of magnitude higher cost. Finally, MMLU evaluation before and after fine-tuning shows that models trained on the Claude-Opus-4.6-generated data retain essentially all of their general knowledge, in contrast to a small but measurable forgetting effect for models trained on the Qwen-generated data. Together, these datasets and the pipeline behind them offer a practical, reproducible path toward scalable industrial benchmarks and training data derived from real-world documentation.
\end{abstract}


\vspace{1em}

\section{Introduction}
In today's industrial world, documents and industrial reports play a key role as vital and specialized knowledge repositories. This knowledge has been gathered over decades and is considered a valuable resource. These documents include a wide range of data such as technical information, product catalogs, maintenance manuals, and specialized guides, all of which carry deep expertise in their respective fields.\cite{bakhtiari2024knowledge} In the "Industry 4.0 framework," such documentation forms the foundation for critical applications like condition-based maintenance and Failure Modes and Effects Analysis (FMEA). For example, ISO standard documents and reports compiled by professionals provide a mapping between potential failures of industrial assets and sensor parameters used to detect faults and anomalies\cite{constantinides2025failuresensoriqmultichoiceqadataset}.

Despite the high value of the information embedded in these reports, organizations face significant challenges in managing and utilizing them. The structure of these documents is often complex and heterogeneous, integrating quantitative data, text, charts, and diagrams seamlessly. Additionally, the knowledge contained in these documents is usually scattered and decentralized; related information about a particular concept or process is distributed across different sections and formats (such as descriptive text, tables, and charts). This makes it difficult to categorize and effectively retrieve information. Traditional information retrieval systems may be able to retrieve relevant data, but they often cannot extract complex patterns and relationships between them.

Tabular data constitutes a significant portion of the specialized knowledge found in industrial and technical documentation. In datasets related to information and communication technology, the text contained in tables accounts for about 18\% of the total word count. For example, in a 6-gigabyte dataset, tables contained 178 million words\cite{min2024exploringimpacttabletotextmethods}. In industry, these tables are compiled by experts and contain very useful information.

These tables are generally valuable because:
\begin{enumerate}
    \item Tables often appear alongside text as they are vital for a comprehensive understanding of the content, such as technical specifications.
    \item Knowledge in these structures directly supports decision-making, risk management, and efficiency enhancement.
    \item By leveraging large language models for reasoning over data and tabular mapping, it becomes possible to simultaneously analyze both the structure of the table and the meaning of its content. These models can automatically extract conceptual relationships and complex dependencies between variables by combining information in the rows and columns; for example, identifying links between different equipment failure modes and patterns recorded in sensor data, or vice versa.
\end{enumerate}

The emergence of large language models, generative AI, and agents has shifted the knowledge management paradigm towards automated extraction and reasoning. Although large language models have been trained on broad and general datasets, they often lack the specialized expertise required for highly technical industrial reports. To bridge this gap, researchers have proposed methods such as fine-tuning and retrieval-augmented generation~\cite{lewis2020retrieval} to inject specialized knowledge hidden in technical manuals and industry-specific data into the models. In this approach, documents are transformed into dynamic knowledge bases, thereby boosting operational effectiveness\cite{bakhtiari2024knowledge,constantinides2025failuresensoriqmultichoiceqadataset}. Results show that the industrial domain is particularly challenging for language models and requires a form of specialized reasoning that general datasets do not encompass. Leading models such as GPT-4\cite{jaech2024openai} and LLama\cite{touvron2023llama} have achieved, on average, only 53.5\% accuracy in question answering on the FailureSensorIQ benchmark with a single correct answer, highlighting the difficulty of industrial data. Open-weight models that were fine-tuned on general datasets such as HotpotQA\cite{yang2018hotpotqadatasetdiverseexplainable} have achieved, on average, 29\% accuracy on industrial benchmarks\cite{constantinides2025failuresensoriqmultichoiceqadataset}.

Overall, large language models, by demonstrating capabilities in both general and specialized fields like medicine, have the capacity for improvement and development in the field of industrial knowledge as well. However, one of the main challenges in this field is the shortage of suitable datasets for training and evaluating these models. The industrial gap refers to the disconnect between the broad, general capabilities of large language models and the highly specialized and precise requirements of the industry. For example, in the semiconductor manufacturing industry, this gap appears as follows\cite{nguyen2024semikongcuratingtrainingevaluating}:
\begin{enumerate}
    \item Lack of deep knowledge: Large language models are trained on massive volumes of textual data and focus on expanding their coverage rather than depth of knowledge.
    \item Mismatch in communication style: Standard models are often designed for general conversations, which creates problems when specialists communicate with each other. In industrial settings, experts do not need general, simple, or summarized explanations; instead, they require precise, concise, and actionable guidance—similar to what an experienced engineer would provide to another experienced engineer. General-purpose models are usually unable to deliver the level of practical applicability and immediate usability needed in real-world production environments.
    \item Inefficiency of standard evaluation criteria: Standard criteria are not suitable for industrial domains and there is a need for specialized datasets and benchmarks.
\end{enumerate}

Despite significant advances in the application of large language models in general domains and some specialized fields, studies show that the volume of research conducted on leveraging real industrial documents—especially focusing on tabular data and the scattered technical knowledge in the documentation of industrial companies—remains limited. Moreover, the majority of recent studies have concentrated on very large language models such as ChatGPT and their counterparts, while smaller models with fewer than ten billion parameters have received less attention. Yet, such smaller models are of great importance in terms of computational cost and deployability in industrial environments.

The present study contributes primarily two datasets, and secondarily the reproducible pipeline that builds them, to bridge the gap between small-scale language models and industrial knowledge. Concretely, our contributions are:
\begin{enumerate}
    \item \textbf{Two industrial QA datasets.} Building on publicly available Panasonic Corporation documents, we design and release the ``Panasonic Dataset'' in two parallel versions---one generated with an open-weight model (Qwen3-30B-A3B-Instruct) and one with a frontier API model (Claude-Opus-4.6)---each pairing instruction-tuning data with a held-out benchmark split for training and evaluating small ($<$10B-parameter) models on real industrial content, including the tabular knowledge that dominates such reports.
    \item \textbf{An end-to-end construction pipeline.} We adapt a multi-scenario retrieval-augmented generation paradigm to real industrial PDFs, combining layout-aware extraction, semantic indexing, and automated quality filtering; while the generation paradigm builds on prior work, applying it to heterogeneous industrial documentation and releasing the resulting artifacts is, to our knowledge, novel.
    \item \textbf{An open- versus frontier-model study.} Producing both dataset versions with the same pipeline lets us compare open-weight and closed API models as data generators along raw data quality, downstream fine-tuning gains, cost, and general-knowledge retention (MMLU).
\end{enumerate}
The raw source corpus contains 906 documents across 7,525 pages, manually curated and downloaded from Panasonic's website.

Our code, dataset generation pipeline, and evaluation scripts are publicly available at \url{https://github.com/parssky/industrial-instruction}. The two released datasets are archived on Hugging Face at \url{https://huggingface.co/datasets/Parssky/industrial-instruction-dataset} (DOI: 10.57967/hf/10098)~\cite{parsa_bakhtiari_2026}.

\section{State of the Art}
One of the important studies in the field of industrial large language models is the FailureSensorIQ research\cite{constantinides2025failuresensoriqmultichoiceqadataset}. It is a specialized evaluation system based on multiple-choice questions and answers. This framework is designed to assess the capabilities of large language models in the field of "Industry 4.0" and focuses on the complex relationships between failure modes and sensor data in various industrial assets. This dataset includes 8,296 questions selected by experts, representing 10 distinct assets, including electric motors, steam turbines, and power converters. These multiple-choice questions were generated in an automated pipeline using ISO document data.
In this study, two main tasks are addressed:
\begin{enumerate}
    \item Failure Mode to Sensor: Identifying the most relevant sensors for detecting the early signs of a specific failure.
    \item Sensor to Failure Mode: Determining which failure mode is likely based on abnormal sensor data.
\end{enumerate}
The results of this evaluation indicate a high level of difficulty in this domain , with the best models achieving an average accuracy of 53\%. This dataset also showed that  stress and variations in the questions can reduce model accuracy to as low as 12\%. In this study, a feature selection tool based on large  language models was also introduced. This tool uses large language models to suggest relevant sensor variables for predictive modeling . This could be a paradigm shift in "Industry 4.0" where large language models act as knowledge producers to assist engineers in feature engineering and root cause failure analysis. This research shows that prompts focused on reasoning  improve the performance of medium-sized models, while agent-based systems  relying on external knowledge —which use knowledge bases such as Wikipedia— often cannot provide the same level of reliability.
Overall, this benchmark determines whether the internal logic of large language models can withstand the noise and complexities of a real factory without getting confused by irrelevant data or minor changes in how a problem is phrased\cite{constantinides2025failuresensoriqmultichoiceqadataset}.
Another challenge in industrial data is the management of mixed data. Data that includes both text and structured and semi-structured tables. There are common approaches for processing this type of data , such as flattening  tables, which causes the loss of structure , or mapping text and tables to separate vector spaces  , which disrupts their semantic link. D. Min et al~\cite{min2024exploringimpacttabletotextmethods}
addresses this gap by integrating the table-to-text conversion process to enhance a question-and-answer system based on large language models. This approach transforms structured tables into coherent natural language descriptions to preserve informational links and structural integrity. In this work, four main categories of table-to-text conversion methods are compared in two main industrial AI paradigms: domain-specific fine-tuning  and retrieval-augmented generation:
\begin{itemize}
    \item Markdown template: a simple, syntax-based method that renders tables in Markdown format.
    \item Template-based Serialization: Uses hand-designed templates based on table features to generate descriptions. 
    \item Traditional Pre-trained Language Model: This method uses pre-trained traditional models like BART to generate text from tables.
    \item Large Language Model-Based Methods: In this method, large language models such as ChatGPT are used to describe the table.
\end{itemize}

The experimental results show that the choice of the table-to-text conversion method has a significant impact on system performance, with differences in various metrics ranging from 2.8\% to 16\% \cite{min2024exploringimpacttabletotextmethods}.
In the domain-specific fine-tuning paradigm, methods based on large language models and traditional pre-trained models have been shown to outperform others, as they generate a higher frequency of specialized terminology and diverse verbs—elements that are critical for acquiring genuine industrial knowledge during the pre-training process.In contrast, in the retrieval-augmented generation paradigm, while methods based on large language models remain suitable for table processing, the Markdown format has also demonstrated high effectiveness\cite{min2024exploringimpacttabletotextmethods}. The reason is that Markdown provides a semantic representation compatible with retrieval, allowing query vectors to embed  and achieve better  alignment with different parts of the documents. Researchers propose language model–based and Markdown–based strategies for industrial data. Consequently, for large language models to play an effective role in industry, they must adopt strategies that bridge the gap between technical data and natural language–based reasoning\cite{min2024exploringimpacttabletotextmethods}.
In a study by H. Femmer et al.\cite{femmer2025descriptioncomparativeanalysisqure} in the field of industrial requirements engineering and the evaluation of large language models, the QuRE dataset comprises 2,111 built-industry requirements gathered from real automotive projects at Mercedes-Benz, which were labeled over a decade through an industrial review process. Unlike other datasets, this dataset provides a gold standard for the quality of natural language requirements and specifically focuses on weak words  —words that may indicate ambiguity, inconsistency, or incompleteness. This research is significant for the industry because it enables a comparative analysis between real-world requirements and requirements generated by language models . QuRE research indicates that requirements generated by models like GPT-4 often exhibit a lower degree of realism compared to real-world industrial data. Synthetic requirements have been found to be syntactically simpler, with shorter sentences, fewer paragraphs, and shallower syntactic trees compared to industrial samples. Although the text generated by large language models is more diverse, it lacks structural complexity and nuance. Real requirements are typically less readable compared to the simple, book-like examples generated by large language models. QuRE studies show that engineering prompt methods  , such as content and  role, also introduce only a limited increase in requirement complexity, but the outputs still have a significant gap in complexity compared to the real industrial dataset. This dataset provides a realistic foundation for evaluating automated quality assessment tools. Because it includes human labeling of a word's content weakness, it turns the dataset into a rigorous test for evaluating the context-based reasoning of language models. Using QuRE, researchers can answer the fundamental question of whether large language models can be used ly to generate high-quality, realistic industrial requirements that reflect the depth and structure of documents used in sensitive and high-risk domains such as automotive, aviation, and healthcare.

\section{Methodology}
Considering the research conducted in the industrial sector, this study aims to fill the gap of insufficient datasets for training and evaluating large language models using an information retrieval-augmented generation approach. In this approach, we seek to create an industrial dataset in the form of multiple-choice question-and-answer pairs and their associated documents, enabling optimal model evaluation and training. To this end, we collected documents related to Panasonic that are publicly available in PDF format and ultimately succeeded in generating 13,557 question-and-answer pairs.
\subsection{Information extraction methods}
Extracting data and information from PDF files is considered a critical bottleneck, especially in retrieval-augmented generation tasks used by large language models. Unlike formats like HTML, the PDF format typically only stores character placement instructions, which makes it challenging to preserve document layout, paragraph coherence and continuity, and word order during information extraction. Existing methods can be divided into two categories: rule-based  and learning-based  , each offering varying levels of capabilities such as Markdown output, image handling, and model compatibility.
\begin{itemize}
    \item Rule-based methods for PDF processing: In these methods, algorithms are used to identify document elements. These tools are typically suitable for digital documents because they are computationally fast and easy to implement. Tools like PyMuPDF are widely used for basic text retrieval. Other tools, such as Camelot, also focus on tabular data  . However, rule-based tools typically perform poorly when dealing with complex, nested tables. Libraries like PyPDF are also capable of extracting and managing images used in documents. Rule-based methods generally fail to preserve semantic structures, which can lead to text fragmentation and reduced performance in information retrieval\cite{Atagong2025A,Nadeem2024Extraction,Bachyr2026Empirical, smith_tesseract_2007}.
    \item Learning-based methods and vision-language models: In this approach, s leverage machine learning and deep neural network architectures such as transformers, especially for complex documents that include images and tables. The transformer-based Nougat~\cite{blecher2023nougat} model is specifically designed to convert scientific documents into Markdown text. This model excels at analyzing mathematical equations and complex formulas. Its Markdown output is very useful for language models, as it preserves the semantic relationship between text and complex elements. The Table Transformer~\cite{smock2022pubtables} model is an object detection model trained on large datasets like PubTables-1M. This model is more flexible in identifying types of tables, such as scientific and financial, compared to rule-based methods. Vision-language models like Dots.OCR and Florance-2 offer an integrated paradigm where layout recognition, optical character recognition, and relationship understanding are learned simultaneously. These models are designed to handle  spatial hierarchies and  semantic granularity, and consequently provide a more robust solution for interpreting structured knowledge from complex visual documents\cite{li2025dotsocrmultilingualdocumentlayout, xiao2023florence2advancingunifiedrepresentation}.
\end{itemize}

\begin{table}[t]
\caption{Comparison between Rule-based and Learning-oriented approaches}
\label{tab:comparison}
\centering
\begin{tabular}{p{3cm} p{4cm} p{4cm}}
\toprule
\textbf{Feature} & \textbf{Rule-based} & \textbf{Learning-oriented} \\
\midrule

Application in language models 
& Suitable for digital government texts; may be fragmented for complex documents.
& Suitable for preserving content; can be converted into a Markdown document. \\

Markdown support 
& Limited; usually requires post-processing after text extraction.
& Native support; designed for outputs like Markdown/LaTeX. \\

Technical images 
& Can extract images but has no relation to the content.
& Ability to recognize objects and link them appropriately to the content. \\

Reliability 
& Highly interpretable; without producing hallucinatory errors.
& Higher risk of hallucinations in complex documents. \\

\bottomrule
\end{tabular}
\end{table}

As shown in Table~\ref{tab:comparison}, for natural language model applications, using learning-oriented methods may be the best option, since they support Markdown, which is one of the bset ways to extrac tables\cite{min2024exploringimpacttabletotextmethods}. Also, since technical documents used in industry include various types of images and tables, many of these documents are stored as scanned or unstructured formats and lack digital layers. Consequently, structured information extraction and maintaining the semantic relationship between text, tables, and images in these documents requires more advanced learning-based methods.
\subsection{Text extraction pipeline from PDF files}
\begin{figure}[t]
    \centering
    \includegraphics[width=1.0\linewidth]{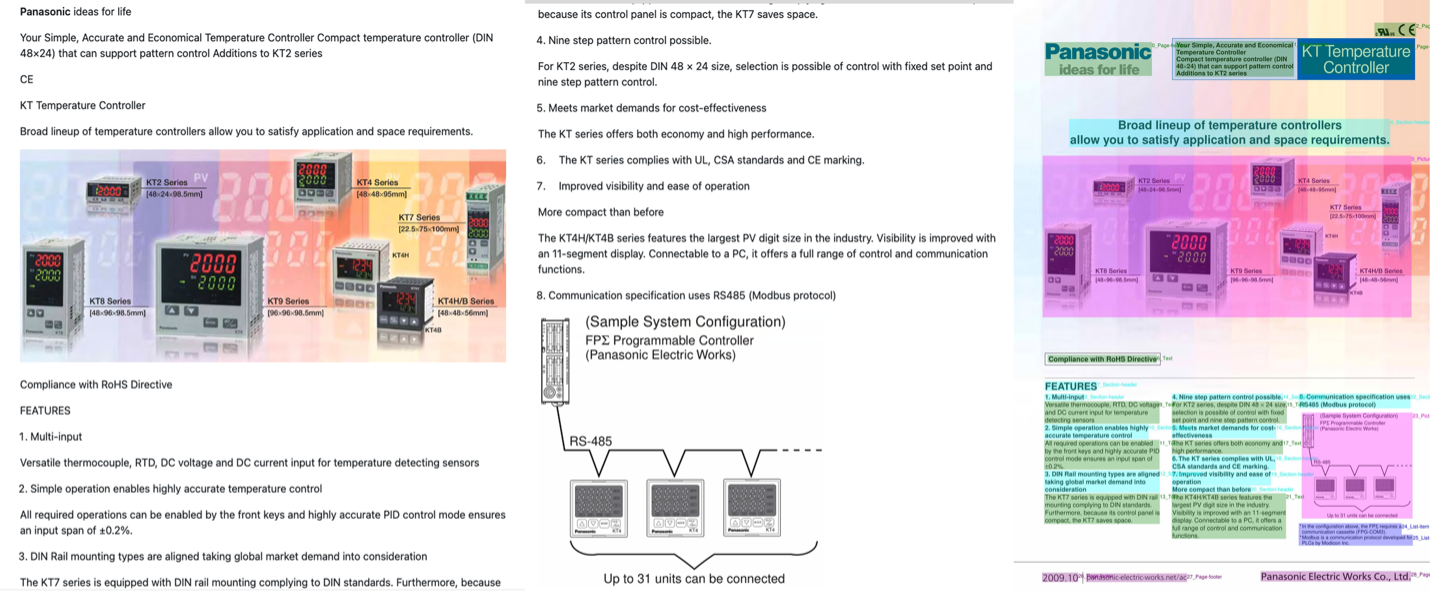}
    \caption{An example output of a page from PDF documents by the Dots.OCR model; on the right, you can see the JPG document that has identified and categorized the various regions and sections, and the two images below are a sample created with the Markdown format, which has placed the images and various sections present on the page into this file.}
    \label{fig:dotsocr}
\end{figure}

In this study, we used the Dots.OCR\cite{li2025dotsocrmultilingualdocumentlayout} model to extract text. This model achieved an OveralEdit score of 0.125 on the \\English OmniDocBench\cite{ouyang2025omnidocbenchbenchmarkingdiversepdf} benchmark, which is significantly better than other tools such as MinerU with a score of 0.150 and Mathpix with a score of 0.191. The OveralEdit metric is a normalized error index based on the edit distance at the level of the entire document, simultaneously reflecting text recognition errors, reading order, and layout structure. This model also achieved a score of 0.177 on the XDocParse benchmark in this metric, compared to the multimodal Gemini-2 model5.-pro performed much better with a score of 0.251. Another competitor, MonkeyOCR-3B, performed worse in the XDocParse benchmark with a score of 0.483. On the other hand, the model's 3 billion parameter size was more efficient in terms of accuracy-to-size ratio\cite{li2025dotsocrmultilingualdocumentlayout}.
Like many vision-language models, the Dots.OCR model is also prone to phenomena such as hallucinations and low localization accuracy, especially when processing dense areas of documents. In other words, any inconsistency in sequence generation can lead to a distorted document layout. One of the main reasons for this issue is the use of downsampling methods in the image feature extraction stage, which removes some of the image's spatial details. This removal of spatial information ultimately leads to the loss of precise alignment between document elements in the output sequence. For example, in documents that consisted entirely of tables, the model failed to correctly extract the structure of the cells and the relationships between them.
Initially, we fed each page of the PDF documents to the model to have it directly generate that same page in Markdown. For each page, the following outputs are produced:
\begin{itemize}
    \item Two Markdown documents (one including header/footer and the other without them).
    \item A JPG document, an example of which can be found at Figure~\ref{fig:dotsocr}. (Image of the page the model used for OCR)
    \item A structured JSON document containing an array of extracted regions, the type of each region, and the extracted text.
\end{itemize}

These outputs, which can be viewed at and exemplified in Figure~\ref{fig:dotsocr}, allow for a qualitative review of the image-to-text extraction performance and the identification of missed regions. In the Markdown document, images are also stored in base64 format. Given the model's weaknesses, pages from which text was not extracted must also be identified. In total, out of 906 documents with 7,525 pages, 291 pages had extraction issues, which on average represents 3.8% of the documents. 

\begin{table}[t]
\small
\caption{ This presents the distribution of document sections that were not successfully processed, categorized by the type of structural element responsible for the processing error. The counts indicate the number of pages affected by each category.}
\label{tab:error_distribution}
\centering
\tiny
\begin{tabular}{lccccccc}
\toprule
\textbf{Classification} & \textbf{Table} & \textbf{Text} & \textbf{Footnote} & \textbf{Index} & \textbf{Section} & \textbf{Page} & \textbf{Etc.} \\
\midrule
Count (pages) & 236 & 14 & 6 & 4 & 2 & 1 & 28 \\
\bottomrule
\end{tabular}
\end{table}

\subsection{Knowledge Base Structuring and Preprocessing}
In the data obtained from the previous stage, there are three types of data: images, text, and tables. The extracted tables are presented as text data in Markdown format, and the images are available as Base64-encoded data. However, the primary focus of this research is on text-based language models rather than vision-language models. Accordingly, during the dataset construction process, only text data and tables were retained, and all images were completely removed. The goal of this selection is to evaluate the ability of purely textual language models to understand and process technical documents without relying on visual information. The final dataset is organized in the standard Huggingface format. Additionally, during the initial preprocessing, pages that were empty, had very little content, or were duplicates were removed, which in total resulted in the deletion of 370 pages from the dataset.

\begin{figure}
    \centering
    \includegraphics[width=0.7\linewidth]{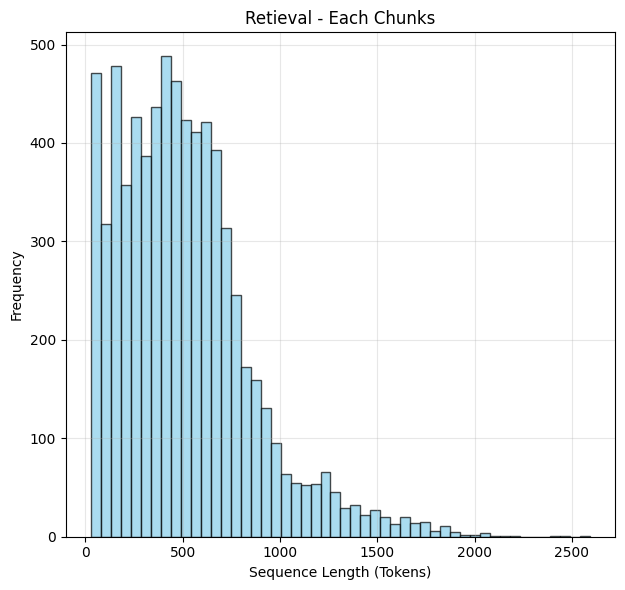}
    \caption{Shows the distribution of tokens across document pages.}
    \label{fig:num_token}
\end{figure}

The combination of the EmbeddingGemma\cite{vera2025embeddinggemma} embedding model with FAISS\cite{Johnson2017} as the retrieval engine creates a highly efficient pipeline for semantic search  and large-scale information retrieval. The EmbeddingGemma model is a model derived from Gemma3~\cite{gemmateam2025gemma3} that has 300 million parameters and, despite its simplicity, performs very powerfully. It also supports  dimension reduction from 768 to 128 without significant loss. This feature allows researchers to balance dimension size, storage space, and search speed. The FAISS library is also a specialized billion-scale search library that enables vector processing. This combination was chosen for its lightness and efficiency in retrieval. Figure~\ref{fig:num_token} shows the distribution of tokens per page. Most documents have around 500 tokens, with longer documents reaching up to 2,500 tokens.  

\subsection{Dataset Creation}
\label{sec:dataset_creation}
\begin{figure}
    \centering
    \includegraphics[width=1.0\linewidth]{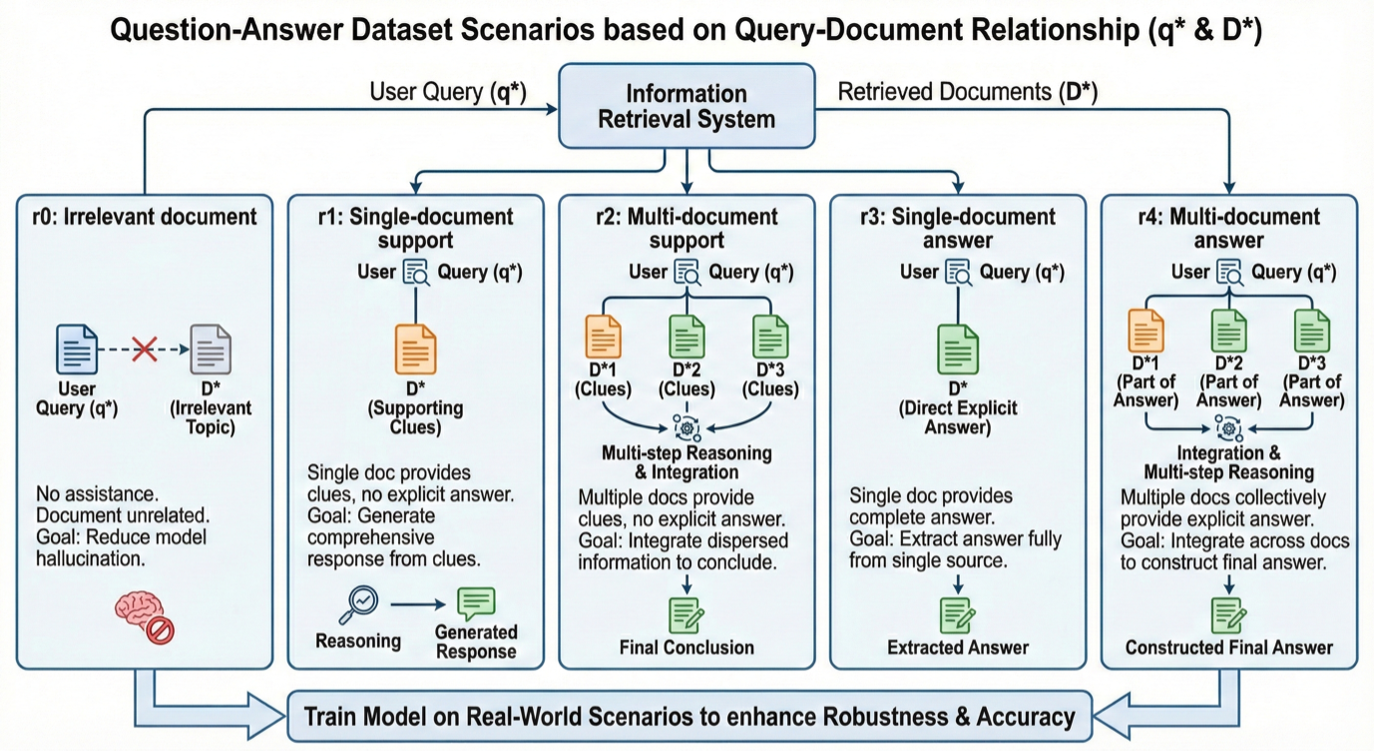}
    \caption{Real-world information retrieval system scenarios and how they are used in an information retrieval-enhanced generation system.}
    \label{fig:scenarios}
\end{figure}

Now, with the information retrieval system in place, we aim to build a question-and-answer dataset to train the model on five real-world scenarios, which you can see in Figure~\ref{fig:scenarios}. These scenarios illustrate the relationship between the user's query q* and the retrieved documents D*:
\begin{enumerate}
    \item Useless Document (r0): In this scenario, the retrieved document provides no help to the model in generating a response; it may even be completely irrelevant. The purpose of training this pattern is for when the model, upon seeing an irrelevant document, begins to hallucinate. Training this pattern helps reduce hallucination in the model.
    \item Single Document Support (r1): In this case, a single document provides supporting information or clues related to the query, but does not contain an explicit and direct answer. The model must use these clues s to generate a comprehensive response.
    \item Multi-Document Support (r2): In this scenario, multiple documents provide clues or supporting information without offering an explicit answer to the query. This mode requires multi-step reasoning or the integration of scattered pieces of information to arrive at a final conclusion.
    \item Single-Document Answer (r3): In this case, a single document provides the complete answer to the question directly. This is the simplest information retrieval scenario, where the answer can be extracted entirely from one source.
    \item Multi-document Answer (r4): In this scenario, multiple documents collectively provide the question's explicit answer. Similar to the multi-document support case, this mode also requires multi-step reasoning or the integration of information across several documents to construct the final answer.
\end{enumerate}

To generate a dataset based on these scenarios, we can use multimodal large language models. This approach offers several advantages that lead to improved small models. Multimodal large language models have a strong ability to produce accurate, high-quality, and instruction-aligned responses, which can serve as a reliable source for training other models. Leveraging its ability to understand these same external source documents, a large language model can accurately and purposefully generate retrieval-aware relationships between a query and a document, for example, by recognizing that a document is intentionally useless~\cite{wang2023selfinstruct}. We also use an instruction simulation method to avoid repetitive instructions. In the prompt we give the model to generate the question and answer, we use similar instructions. This method ensures that the resources the model references are not repetitive and include various forms of instructions.
\\
Now, selecting a model for the instruction-following task is important. Large, multi-purpose language models like GPT-4o can be used, but alternatives can also be found. In the context of synthetic dataset generation, the Qwen3 model, particularly the Qwen3-30B-A3B-Instruct, serves as a lightweight alternative to GPT-4o due to its combinatorial reasoning capabilities, high instruction-following accuracy, and extensive multilingual coverage. The following points can be noted when comparing these two models\cite{yang2025qwen3}:
\begin{enumerate}
    \item Superior adherence to instructions and alignment: dataset generation requires strict adherence to formatting and complex logic. Benchmarks show that Qwen3-30B-A3B-Instruct scores 84.7 in IFEval (instruction following) achieved a score of 84.7, which is higher than the 83.9 score for GPT-4o. Additionally, this model reached a score of 69 on Arena-Hard-V2 (human-aligned), a benchmark for instruction-tuned models, which is better than the 61.9 recorded for GPT-4o. Also, on the WritingBench benchmark (Creative Writing) the Qwen model has reached a score of 85.5, which is higher than the 75.5 for GPT-4o.
    \item Advanced Reasoning Through Hybrid Modes: Unlike traditional large models, Qwen3 introduces a hybrid reasoning framework that switches between "Thinking" and "Non-Thinking" modes. The "thinking" mode is specifically optimized for multi-step logical reasoning, mathematics, and programming, while the "non-thinking" mode is suitable for fast, general, and creative tasks.
\end{enumerate}

Given its superior performance in evaluations and its open-source nature, the Qwen3-30B-A3B-Instruct model was used to generate the dataset. As shown in Figure~\ref{fig:scenarios} the various sections are separated by markers to help the model effectively identify the documents and random instructions.

\begin{lstlisting}[style=promptstyle, float=tbp, caption={The prompt sample used for scenario r0, which is placed in the first part of the retrieved documents, followed by a random instruction in the next section to ensure the generated samples are appropriate.}, label={lst:prompt_sample}]
<Documents>
{doc}
</Documents>
Your task is to generate an English question q* and a corresponding response a* based on the provided <Documents>. Please note that the question q* can take various forms, not limited to questions with a question mark, but also including statements, instructions, and other formats. You need to follow the requirements below to generate the q* and a* (RAG Paradigms):
1. q* should be related to the <Documents>, but the <Documents> can not provide any useful information for answering q*.
2. a* should be able to answer q*, ensuring that the response a* is accurate, detailed, and comprehensive.
3. q* must be a standalone question. Strictly avoid using phrases that reference the source material, such as "Based on the provided context," "According to the documents," "In the text," or similar meta-references.
Additionally, to ensure diversity, richness, and high quality in the question q* you generate, we will randomly provide a question for you to emulate. In other words, while satisfying the requirements above, make q* similar in task requirement and expression to the <Simulated Instruction> below:
<Simulated Instruction>
{simulated_instruction}
</Simulated Instruction>
IMPORTANT: If the <Simulated Instruction> is a multiple-choice question with options, then your generated q* MUST ALSO include exactly 5 options labeled A-E, and a* MUST include the correct option(s) in the same JSON list format used in the simulated instruction. Always generate 5 options, even if the original document does not contain any option-like content.
Please directly generate the question-answer-options (q*, a*, options*) following all the rules above in the format of {"q*": ..., "a*": ..., "options*": ...}, so for a* select A-E correct option.
Ensure the quality of the generated (q*, a*, options*).
\end{lstlisting}

To simulate multiple-choice questions that also allow for accuracy measurement, we used the SensorIQ dataset instructions from IBM. This dataset includes 5,334 multiple-choice questions and answers that are used for each of the five scenarios. As shown in the Figure~\ref{lst:prompt_sample} , the dataset generation pipeline first selects one of the SensorIQ  datasets, then, based on the prompt, retrieves three documents from the knowledge base and feeds them to the large Qwen model to generate the corresponding question and answer for the scenario. 

\begin{figure}
    \centering
    \includegraphics[width=1.0\linewidth]{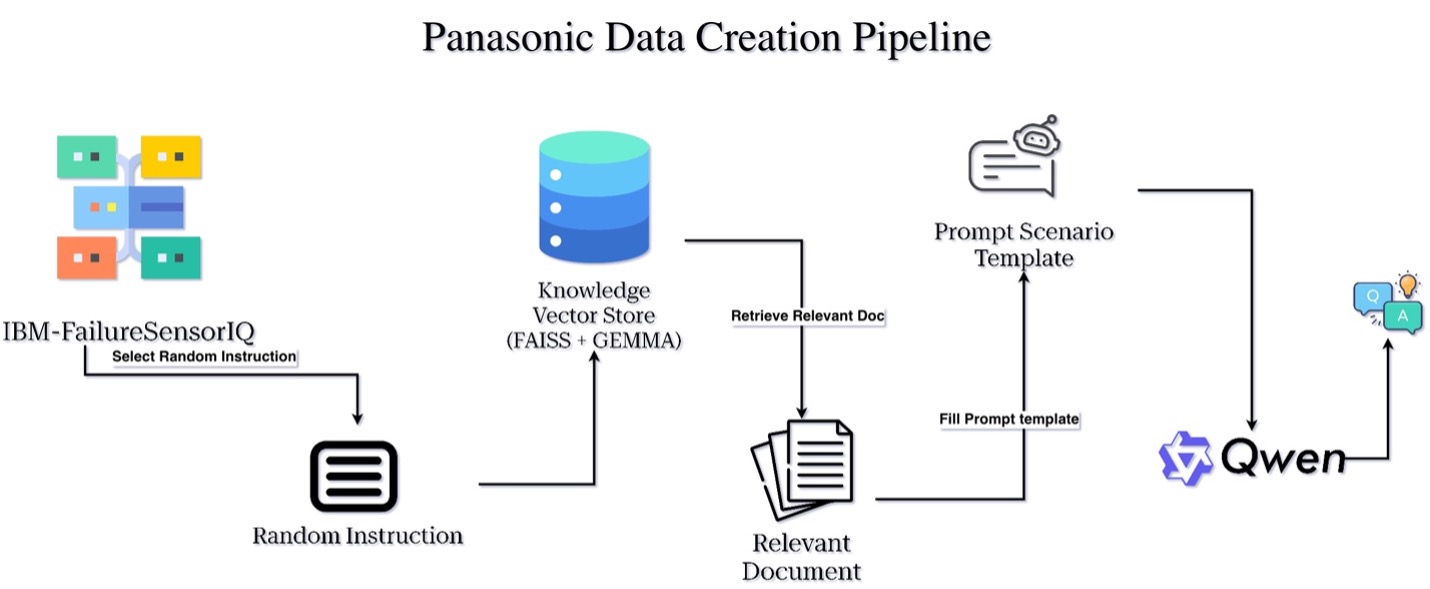}
    \caption{The pipeline for generating the Panasonic document dataset.}
    \label{fig:pipeline}
\end{figure}

Thus, a total of 23,910 Q\&As are generated. During sample generation, some samples were dropped by Qwen because, due to the probabilistic nature of large language models' outputs, they did not follow the specified JSON structure. In Figure~\ref{fig:pipeline}, this structure is in JSON format so that each section can be examined with code.

\begin{lstlisting}[style=promptstyle, float=tbp, caption={A sample generated by Qwen that includes the question, answer, options, and related documents.}, label={lst:generated_sample}]
{
    "q*": "When selecting a current sensing resistor for use in a high-reliability automotive application, which performance characteristic should be prioritized to ensure long-term stability under thermal cycling conditions?",
    "a*": ["A"],
    "options*": ["A Thermal shock resistance",
                 "B Solderability",
                 "C Frequency characteristics",
                 "D Moisture resistance",
                 "E Overload capability"],
    "documents": ["Current Sensing Resistors, Metal Plate Type\n\n## Performance (AEC-Q200) \n\n### ERJMS4S/ERJMS4H\n\n ... "]
}
\end{lstlisting}

\subsubsection{Question and Answer Preprocessing}

\textbf{Filter for detecting valid answer options.} To enhance the quality of the input data and ensure the validity of the multiple-choice question structure, a rule-based preprocessing has been employed. This preprocessing is tasked with selecting samples whose options have a complete and meaningful structure. The goal is to determine whether the options truly consist of an identifier and an explanatory text, or are merely a collection of short, incomplete identifiers, such as single digits. In Listing~\ref{lst:generated_sample}, one can see examples of correct and incorrect options. First, it is checked that the number of options is appropriate; then each option is analyzed individually. Options that consist of only a short identifier such as A or 1 and lack explanatory text are considered incomplete options. Options that begin with an identifier and end with explanatory text, separated by punctuation such as a period, parenthesis, or colon, are considered complete options. An example of a question with no explanation in the options, where only the option letter is written, can be seen in Figure~\ref{lst:correct_options}.\\

\textbf{Filter for identifying options embedded in the question text.} Now, in addition to examining the options' structure in the template, another filter is used to detect answers hidden in the question text. This filter is designed to identify instances where options are directly inserted into the question text. For this filter as well, rule-based methods are used, and the specifies whether a minimum number of answer options are structurally present in the question text, an example of which can be seen in Figure~\ref{lst:correct_options}.\\

Finally, we add the list of first-filter samples to the options set and merge both the first and second filter lists together. This set includes q*, which contains the question along with its options. We remove from the dataset the remaining samples that did not fall into either filter. In Listing~\ref{lst:wrong_qa}, you see an example of this category where, in practice, there are no options and only letters are provided as choices. In Table~\ref{tab:scenario_samples}, the number of samples is specified; with this preprocessing method, we ensure that high-quality samples are retained, as they can have a significant impact on the training and evaluation process. 

\begin{lstlisting}[style=promptstyle, float=tbp, caption={Correct and incorrect examples of options in samples generated during the dataset construction process.}, label={lst:correct_options}]
# Correct
A. text
B) description
l: explanation

# Incorrect
A
l relay model
\end{lstlisting}

\begin{lstlisting}[style=promptstyle, float=tbp, caption={In this sample Q\&A, the full descriptions of the options are provided immediately after the question, and in the option template only the letters are displayed without any explanation, which does not follow the original format.}, label={lst:wrong_qa}]
{
    "q*": "Please select the correct option(s) from the following options given the question: Which sensor type is most suitable for detecting motionless objects and providing information about the direction of movement, in addition to measuring temperature? Options: A PIR (Through Hole) B Grid EYE High Gain Type C Grid EYE Low Gain Type D EKM-B E AMN",
    "a*": ["B", "C"],
    "options*": ["A", "B", "C", "D", "E"],
    "documents": ["..."]
}
\end{lstlisting}

\begin{table}[t]
\small
\caption{Total number of samples after preprocessing in each scenario.}
\label{tab:scenario_samples}
\centering
\tiny
\begin{tabular}{lccccc}
\toprule
\textbf{Filter/Scenario} & \textbf{R0} & \textbf{R1} & \textbf{R2} & \textbf{R3} & \textbf{R4} \\
\midrule
Filter to detect healthy response options 
& 2472 & 2526 & 2406 & 2476 & 2379 \\

Filter for identifying options embedded in the question text 
& 427 & 302 & 410 & 315 & 365 \\

Rest (deleted) 
& 1962 & 1961 & 1994 & 2015 & 2052 \\

Total remaining 
& 2899 & 2828 & 2816 & 2791 & 2744 \\
\bottomrule
\end{tabular}
\end{table}

\textbf{Filter for checking the option response.} Also, in each sample, there is a section titled a* which contains the answer to that question. Sometimes the model used a keyword other than a* for construction, or it used a complex form that depended on the technical identifier . This section must also be a list of characters; in some samples, this list is not in that format, and a different type of rule-based filter was used for this section to ensure the responses are in a standard format.\\

\begin{figure}
    \centering
    \includegraphics[width=1.0\linewidth]{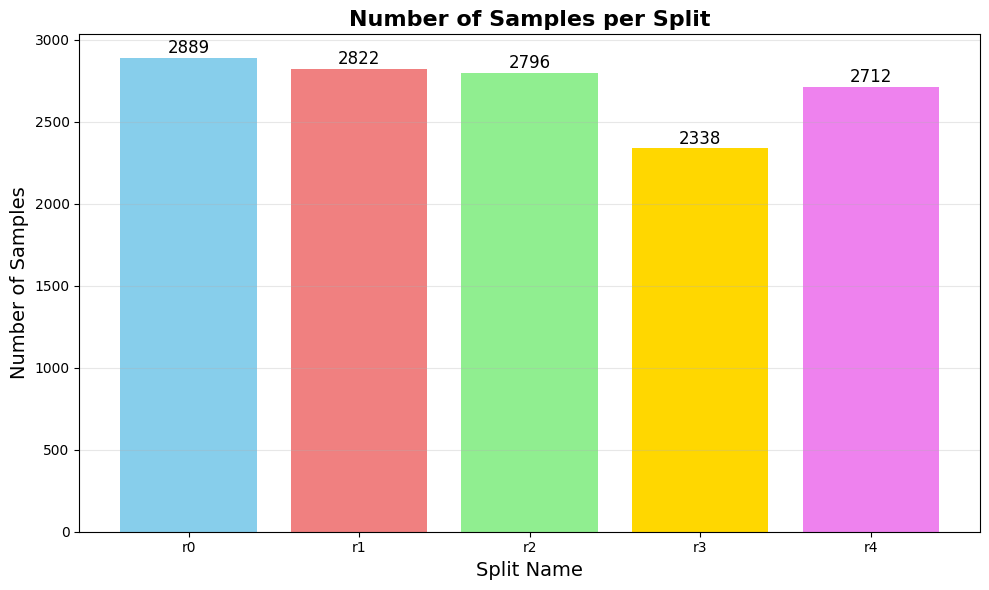}
    \caption{This shows the final set of samples after the filtering process in each scenario.}
    \label{fig:num_of_sample}
\end{figure}

After applying the two initial filters, the option-response filter is also applied, and finally the distribution of the dataset in is shown in Figure~\ref{fig:num_of_sample}. There are also no significant discrepancies in this distribution, and during the model training process it can almost learn the sample proportions. 

At the outset of this study, approximately 24,000 samples were generated, which, after a stringent preprocessing step, were reduced to about 13,500 samples. Given that the behavior and output of language models are a direct reflection of their training data, and that the quality of the dataset plays a decisive role in the model's final performance, we adopted a rigorous approach to data cleansing. Initially, to evaluate and benchmark the models, we randomly selected 200 samples from each defined scenario, the details of which are specified in Table~\ref{tab:data_split}. The remaining samples were then used as the training dataset, with a portion of these data also allocated to the test or validation set during the training process.

\begin{table}[t]
\small
\caption{Data split for training and testing.}
\label{tab:data_split}
\centering
\begin{tabular}{lc}
\toprule
\textbf{Sample Type} & \textbf{Sample Size} \\
\midrule
Training & 12,557 \\
Test (Benchmark) & 1,000 \\
\bottomrule
\end{tabular}
\end{table}
\subsection{Dataset Regeneration with a Frontier API Model}
\label{sec:claude_regeneration}
\begin{figure}
    \centering
    \includegraphics[width=1.0\linewidth]{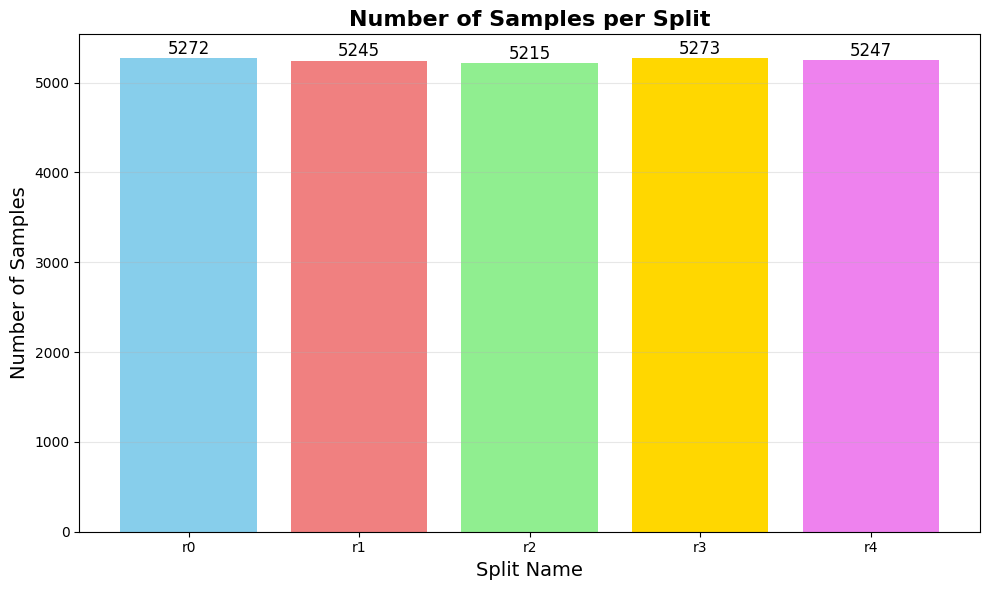}
    \caption{Per-scenario sample counts for the Panasonic dataset generated with Claude-Opus-4.6, after filtering.}
    \label{fig:claude_scenario_dist}
\end{figure}

To test whether the quality of the generator model materially affects the quality of the resulting Industrial-Instruction dataset, we repeated the entire generation pipeline described above using Claude-Opus-4.6 in place of Qwen3-30B-A3B-Instruct, keeping the retrieval pipeline, prompts, and the five scenario definitions (r0--r4) unchanged. This run produced 26,395 raw Q\&A pairs, of which only 143 samples (0.5\%) were removed by the same preprocessing filters described above, compared with roughly 10,353 samples (43\%) removed from the Qwen3-30B-A3B-Instruct output (Table~\ref{tab:scenario_samples}). Figure~\ref{fig:claude_scenario_dist} shows the resulting per-scenario distribution, which remains balanced across the five scenarios (5,215--5,273 samples each). Table~\ref{tab:claude_data_split} reports the resulting training and benchmark splits.

\begin{table}[t]
\small
\caption{Data split for training and benchmarking on the Claude-Opus-4.6-generated dataset.}
\label{tab:claude_data_split}
\centering
\begin{tabular}{lc}
\toprule
\textbf{Sample Type} & \textbf{Sample Size} \\
\midrule
Training & 25,252 \\
Test (Benchmark) & 1,000 \\
\bottomrule
\end{tabular}
\end{table}

The substantially lower filtering rate suggests that Claude-Opus-4.6 followed the required JSON schema and option-formatting constraints more reliably than the open-weight Qwen3-30B-A3B-Instruct model, consistent with its much larger parameter count and correspondingly stronger instruction-following ability. This came at a monetary cost, however: generating the Claude-Opus-4.6 dataset cost approximately \$330 in API usage, versus roughly \$3.2 in local compute for the Qwen3-30B-A3B-Instruct run (Table~\ref{tab:cost_comparison}, Section~\ref{sec:conclusion}).
\section{Experimental Setup}
To examine the extracted dataset, a training and evaluation framework has been set up. First, the models are evaluated on the test portion of the data to obtain the required metrics. Then, the models are trained on the other portion to investigate the dataset's impact on the model. Additionally, in this section, the models are also evaluated on the FailureSensorIQ dataset, and its impact on this data is also determined. 
\subsection{Benchmarking of base models on the test dataset}
To accurately evaluate model performance, selecting an evaluation metric is essential. This metric can vary depending on the task and objective. In the question-answering task, two metrics, Exact-Match and Accuracy, are commonly used.

\begin{itemize}
    \item \textbf{Exact-Match criterion.}A binary criterion that checks whether the predicted string is exactly the same as the reference string, character-by-character. This criterion is very sensitive to reordering and formatting. For example, if the correct answer is [A, B] is the correct answer and the model produces [B, A], the metric assigns a score of zero, even though the answer is logically correct.
    \item \textbf{Accuracy metric.} This metric is typically used for classification and calculates the fraction of samples where the predicted class is the same as the reference class. When applied to lists of string representations, Accuracy behaves similarly to Exact-Match. In this case, the entire string [A, B] is considered as a class label. This metric may also mark a correct sample as incorrect due to differences in order. 
\end{itemize}

Given the limitations of the above metrics in assessing responses that are inherently set-based and order-independent, using composite and set-based metrics provide a more accurate evaluation of the model's performance. Accordingly, the following metrics are used in this study to evaluate the models.

\begin{itemize}
    \item \textbf{Set-Match-Accuracy.}A domain-specific metric that converts string representations into unordered sets before comparison. This metric checks for set equality and is therefore robust to element order and formatting, such as whitespace. This metric aligns the evaluation with the problem's logic and ensures that the model is judged on its ability to identify the correct option, not on a specific order. 
    \item \textbf{F1 Score.} This standard metric for multi-label classification balances accuracy and recall. The accuracy metric measures how many of the options the model selected were correct, while the recall metric measures how many of the correct options available were selected by the model.
    \item \textbf{Jaccard similarity metric.} This metric measures the overlap between the predicted set and the reference set. Its interpretation is intuitive: what percentage of the total unique options involved were correct?
\end{itemize}
Thus, the Set-Match-Accuracy, F1-Score, and Jaccard similarity metrics were chosen as the final evaluation criteria to assess the models' performance based on the logical correctness of the answers rather than on any specific ordering or formatting.\\

The comparisons in Section~\ref{sec:industrial_comparison} additionally evaluate models on the FailureSensorIQ benchmark, using four metrics reported by that benchmark's own evaluation harness: \textbf{AccOrgIBM}, accuracy on FailureSensorIQ's original, unperturbed questions; \textbf{AccPerIBM}, accuracy on the perturbed variant of the same questions, used to test robustness to rephrasing~\cite{constantinides2025failuresensoriqmultichoiceqadataset}; and \textbf{F1-Macro}/\textbf{F1-Micro}, the macro- and micro-averaged F1 scores across FailureSensorIQ's question categories. On the Panasonic benchmark we continue to report Set-Match-Accuracy, F1-Score, and Jaccard Similarity as defined above; in the comparison figures these are labeled \textbf{AccPana}, \textbf{F1Score}, and \textbf{JacSim} respectively for brevity.

Also, language models are sensitive to the length of the generated outputs y, in that if you ask the model for only the correct option, it might select one option, but in an explanatory mode, if the model writes an explanation and specifies the reason for the selection, it might choose a different option. Also, in the desired response, only the option matters, and the preceding explanations are not relevant to the evaluation metric. Therefore, a balancing layer is needed to extract the model's final answer. To address this issue, a normalization layer is applied to the model's output. The goal of this layer is to extract the final answer independently of the model's explanatory details. In this approach, the model is asked to provide the final answer in the form of a standard JSON block. Then, using regular expressions, this block is extracted and selected as the reference answer. In the normalization stage, common inconsistencies in the output of language models, such as the use of single quotation marks instead of double quotation marks, are corrected, and the JSON structure is converted into a processable format. This process ensures that the evaluation metric focuses solely on the model's final answer and is not influenced by explanatory explanations or changes in the output's format.

The FailureSensorIQ study addressed large models\cite{constantinides2025failuresensoriqmultichoiceqadataset}, whereas this research focuses on small language models. The purpose of this selection is to examine models that require fewer hardware resources, making them practical for small companies and research teams with limited computational resources. In this regard, the proposed models and their accuracy have been evaluated on the Panasonic industrial dataset, which can be viewed in Table~\ref{tab:base_model_results}.\\

The base model also considered is Qwen-4B-Instruct, and we will proceed to discuss the training and evaluation process for this model.

\begin{table}[t]
\small
\caption{Performance of base models on the Panasonic industrial dataset. Qwen-4B-Instruct, Phi-3-mini-4k-Instruct~\cite{abdin2024phi3}, RAG-Instruct-Llama3-8B}
\label{tab:base_model_results}
\centering
\begin{tabular}{lccc}
\toprule
\textbf{Model} & \textbf{Set-Match Acc.} & \textbf{F1-Score} & \textbf{Jaccard} \\
\midrule
Qwen-4B-Instruct & 28.5\% & 46.65\% & 41.62\% \\
Phi-3-mini-4k-Instruct & 17.5\% & 31.27\% & 28\% \\
RAG-Instruct-Llama3-8B & 0.70\% & 0.90\% & 0.86\% \\
\bottomrule
\end{tabular}
\end{table}

\subsection{General-Knowledge Evaluation Protocol}
\label{sec:mmlu_protocol}
A well-known risk of fine-tuning a pre-trained large language model on a narrow, domain-specific dataset is \textit{catastrophic forgetting}, in which the model loses part of the general-purpose knowledge acquired during pre-training while adapting to the new data distribution~\cite{luo2025empiricalstudycatastrophicforgetting}. The severity of forgetting depends on several factors, including the fine-tuning method, the nature of the training data, and the training duration; prior work shows that instruction-based fine-tuning tends to induce more forgetting than continued pre-training, that forgetting worsens as training progresses, and that parameter-efficient methods such as LoRA, while typically weaker than full fine-tuning on the target task, tend to better preserve out-of-domain performance~\cite{biderman2024loralearnslessforgets}. Because the degree of forgetting cannot be predicted from the training method alone, it must be measured empirically with standard benchmarks; evaluating general knowledge before and after fine-tuning is a widely adopted protocol in the large-language-model literature~\cite{luo2025empiricalstudycatastrophicforgetting}.

We use the Massive Multitask Language Understanding (MMLU) benchmark~\cite{hendrycks2021measuringmassivemultitasklanguage} to evaluate general knowledge and reasoning ability. MMLU spans 57 subjects across STEM, the humanities, the social sciences, and specialized domains such as law and medicine, with difficulty ranging from introductory to expert level, and is designed to test whether a model has acquired generalizable world knowledge during pre-training. For this reason, MMLU is one of the most commonly used measures of general-knowledge retention in the catastrophic-forgetting literature. In this study we measure accuracy on the full MMLU test set (57 subjects grouped into 4 top-level categories, 14,042 questions) both before and after fine-tuning on the Panasonic industrial dataset, to determine whether fine-tuning degrades the model's ability to answer general-domain questions.

\subsection{Training the Models on the Qwen-Generated Dataset}
\label{sec:qwen_training}
Now, we prepare the base model on the training dataset and measure the impact of that data on the model. Fine-tuning is the process of adapting a pre-trained large language model to a specific task or domain by continuing to train the model on a smaller, domain-specific dataset. Main fine-tuning methods include  , in which all weights are updated, and  , which only modifies a small portion of the weights. Among PEFT methods,LoRA is the most common approach, approximating a full weight update by applying low-rank matrix multiplications. Despite the high efficiency of the LoRA method, it experiences a performance drop compared to full fine-tuning, especially when the target task requires high levels of logical reasoning or learning entirely new skills\cite{hu2022lora, Wang2024Parameter-efficient, Anisuzzaman2024Fine-Tuning, dettmers2023qloraefficientfinetuningquantized}.

The results of the LoRA method do not show significant changes in the model. Four experiments were conducted to improve the Qwen3-4B-Instruct model, and the evaluation results on the Panasonic dataset can be found in Table~\ref{tab:lora_results}.

\begin{table}[t]
\small
\caption{Evaluation results under different LoRA fine-tuning configurations.}
\label{tab:lora_results}
\centering
\begin{tabular}{lccc}
\toprule
\textbf{Config} & \textbf{F1-Score} & \textbf{Jaccard Similarity} & \textbf{Set-Match Acc.} \\
\midrule
Original & 46.57\% & 41.57\% & 28.50\% \\
R16-A16  & 46.71\% & 41.66\% & 28.50\% \\
R32-A16  & 46.89\% & 41.86\% & 28.70\% \\
R8-A16   & 46.53\% & 41.54\% & 28.50\% \\
R64-A16  & 46.27\% & 41.27\% & 28.20\% \\
\bottomrule
\end{tabular}
\end{table}

Also, for a more detailed examination, the error rate chart for is available in Table~\ref{tab:lora_loss} . In this table, the charts are nearly identical, and changing the LoRA-related parameters had no impact on the error rate and, consequently, did not improve the model.

\begin{table}[t]
\normalsize
\caption{Loss reduction under different LoRA rank configurations.}
\label{tab:lora_loss}
\centering
\begin{tabular}{ccc}
\toprule
\textbf{R} & \textbf{Alpha} & \textbf{Loss Changes} \\
\midrule
\multirow{4}{*}{}
8  & 16 & \multirow{4}{*}{\includegraphics[width=0.32\columnwidth]{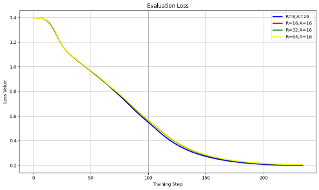}} \\
16 & 16 & \\
32 & 16 & \\
64 & 16 & \\
\bottomrule
\end{tabular}
\end{table}
Therefore, one can say that this method was not effective for training the Qwen model on the dataset. Of course, LoRA fine-tuning is for aligning the model's behavior and cannot affect the model's knowledge. For this reason, we pursued full fine-tuning and trained all of the model's parameters. The results are available at and in Table\ref{tab:finetune_rag_results} . With full fine-tuning, the Set-Match Accuracy metric increased from 28.5\% to 42\%. The F1-Score and Jaccard-Similarity metrics also reached 63.48\% and 57.95\%, respectively, which are better than the baseline results in Table~\ref{tab:base_model_results}.

\subsection{Training the Models on the Claude-Generated Dataset}
\label{sec:claude_training}
Following the same full-fine-tuning protocol described above — identical hyperparameters, optimizer settings, and training duration — we fine-tuned Qwen3-4B-Instruct on the Claude-Opus-4.6-generated Panasonic dataset (Table~\ref{tab:claude_data_split}) to allow a controlled comparison between the two data sources. Table~\ref{tab:claude_finetune_results} reports the results on the corresponding held-out test split.

\begin{table}[t]
\small
\caption{Performance comparison before and after full fine-tuning, with and without RAG, on the Claude-Opus-4.6-generated Panasonic dataset.}
\label{tab:claude_finetune_results}
\centering
\scriptsize
\begin{tabular}{lccc}
\toprule
\textbf{Configuration} & \textbf{F1-Score} & \textbf{Jaccard Similarity} & \textbf{Set-Match Acc.} \\
\midrule
Original with RAG & 58.55\% & 54.00\% & 40.90\% \\
Fully Fine-Tuned with RAG & \textbf{72.66\%} & \textbf{68.88\%} & \textbf{56.40\%} \\
\midrule
Original & 59.24\% & 54.71\% & 41.60\% \\
Fully Fine-Tuned & 72.72\% & 68.93\% & 56.40\% \\
\bottomrule
\end{tabular}
\end{table}

Fine-tuning on the Claude-Opus-4.6-generated data produces larger absolute gains than fine-tuning on the Qwen3-30B-A3B-Instruct-generated data (Table~\ref{tab:finetune_rag_results}): Set-Match Accuracy improves by roughly 15 percentage points with RAG (40.9\%$\rightarrow$56.4\%), versus about 13.5 points for the Qwen-generated dataset (28.5\%$\rightarrow$42.0\%). Note, however, that the two evaluations use different held-out test splits (each generated by its respective model), so the ``Original'' baseline scores also differ between Table~\ref{tab:finetune_rag_results} and Table~\ref{tab:claude_finetune_results} (28.5\% vs. 40.9--41.6\%); the two improvement magnitudes are therefore indicative rather than strictly controlled, and should not be over-interpreted as a head-to-head comparison.

\begin{table}[t]
\small
\caption{Performance comparison before and after full fine-tuning, with and without RAG, on the Panasonic dataset.}
\label{tab:finetune_rag_results}
\centering
\scriptsize
\begin{tabular}{lccc}
\toprule
\textbf{Configuration} & \textbf{F1-Score} & \textbf{Jaccard Similarity} & \textbf{Set-Match Acc.} \\
\midrule
Original with RAG & 46.57\% & 41.57\% & 28.50\% \\
Fully Fine-Tuned with RAG & \textbf{63.48\%} & \textbf{57.95\%} & \textbf{42.00\%} \\
\midrule
Original & 46.63\% & 41.61\% & 28.50\% \\
Fully Fine-Tuned & 63.14\% & 57.60\% & 41.70\% \\
\bottomrule
\end{tabular}
\end{table}

\section{Results and Discussion}
This study uses two benchmarks to evaluate model performance: FailureSensorIQ and the Panasonic industrial dataset introduced above. In all evaluation scenarios, the retrieval component draws from the Panasonic knowledge base described in Section~\ref{sec:claude_regeneration}. We first examine the effect of fine-tuning on each model's general knowledge (Section~\ref{sec:mmlu_results}), then compare the base Qwen3-4B-Instruct model, its Panasonic-fine-tuned variants, and RAG-Instruct-Llama3-8B across both benchmarks for each of the two dataset-generation approaches used in this study (Section~\ref{sec:industrial_comparison}).

\subsection{General-Knowledge Evaluation Results}
\label{sec:mmlu_results}
Table~\ref{tab:mmlu_overall} reports overall MMLU accuracy for the base model (Qwen3-4B-Instruct-2507, no fine-tuning), the model fine-tuned on the Qwen3-30B-A3B-Instruct-generated dataset (\texttt{finetuned\_v1\_qwen}), and the model fine-tuned on the Claude-Opus-4.6-generated dataset (\texttt{finetuned\_v2\_claude}).

\begin{table}[t]
\small
\caption{MMLU accuracy of the base model and the two fine-tuned variants.}
\label{tab:mmlu_overall}
\centering
\begin{tabular}{lc}
\toprule
\textbf{Model} & \textbf{MMLU Overall} \\
\midrule
Original (base) & 72.13\% \\
finetuned\_v1\_qwen & 70.87\% \\
finetuned\_v2\_claude & 72.08\% \\
\bottomrule
\end{tabular}
\end{table}

\texttt{finetuned\_v2\_claude} retains essentially all of the base model's general knowledge (a 0.05-point drop), whereas \texttt{finetuned\_v1\_qwen} loses 1.26 points overall. Neither drop is severe on its own, but a clear pattern emerges once results are broken down by top-level category (Table~\ref{tab:mmlu_category}).

\begin{table}[t]
\small
\caption{MMLU accuracy by top-level category for the base model and the two fine-tuned variants.}
\label{tab:mmlu_category}
\centering
\begin{tabular}{lccc}
\toprule
\textbf{Category} & \textbf{Original} & \textbf{finetuned\_v1\_qwen} & \textbf{finetuned\_v2\_claude} \\
\midrule
Humanities & 63.66\% & 61.13\% & 63.85\% \\
Social Sciences & 81.38\% & 80.44\% & 81.28\% \\
STEM & 72.53\% & 72.41\% & 72.44\% \\
Other & 75.41\% & 74.57\% & 75.09\% \\
\bottomrule
\end{tabular}
\end{table}

\texttt{finetuned\_v2\_claude} performs on par with, or marginally better than, the base model in every category, with only a small dip in the ``Other'' category — one of the clearest cases of negligible catastrophic forgetting observed in this study. \texttt{finetuned\_v1\_qwen}, by contrast, shows forgetting concentrated in the Humanities category, driven in particular by a roughly 10-point drop on moral-reasoning-related subjects (see Appendix~\ref{app:mmlu_detail} for the subject-level breakdown). \texttt{finetuned\_v2\_claude} additionally shows small but consistent \emph{improvements} over the base model on several quantitative- and legal-reasoning subjects, including \texttt{global\_facts}, \texttt{college\_mathematics}, \texttt{machine\_learning}, \texttt{security\_studies}, and \texttt{international\_law}. This indicates that fine-tuning on the higher-quality Claude-Opus-4.6-generated dataset can, in this case, improve certain general-reasoning capabilities without sacrificing others — a much more favorable trade-off than the one observed for \texttt{finetuned\_v1\_qwen}.

\subsection{Comparison on Industrial Benchmarks}
\label{sec:industrial_comparison}
In this section we compare each fine-tuned model against both the FailureSensorIQ and Panasonic benchmarks. In all cases, the retrieval component is implemented using the Panasonic knowledge base described in Section~\ref{sec:claude_regeneration}.

\subsubsection{Qwen-Generated Dataset}
\label{sec:comparison_qwen}
\begin{figure}
    \centering
    \includegraphics[width=1.0\linewidth]{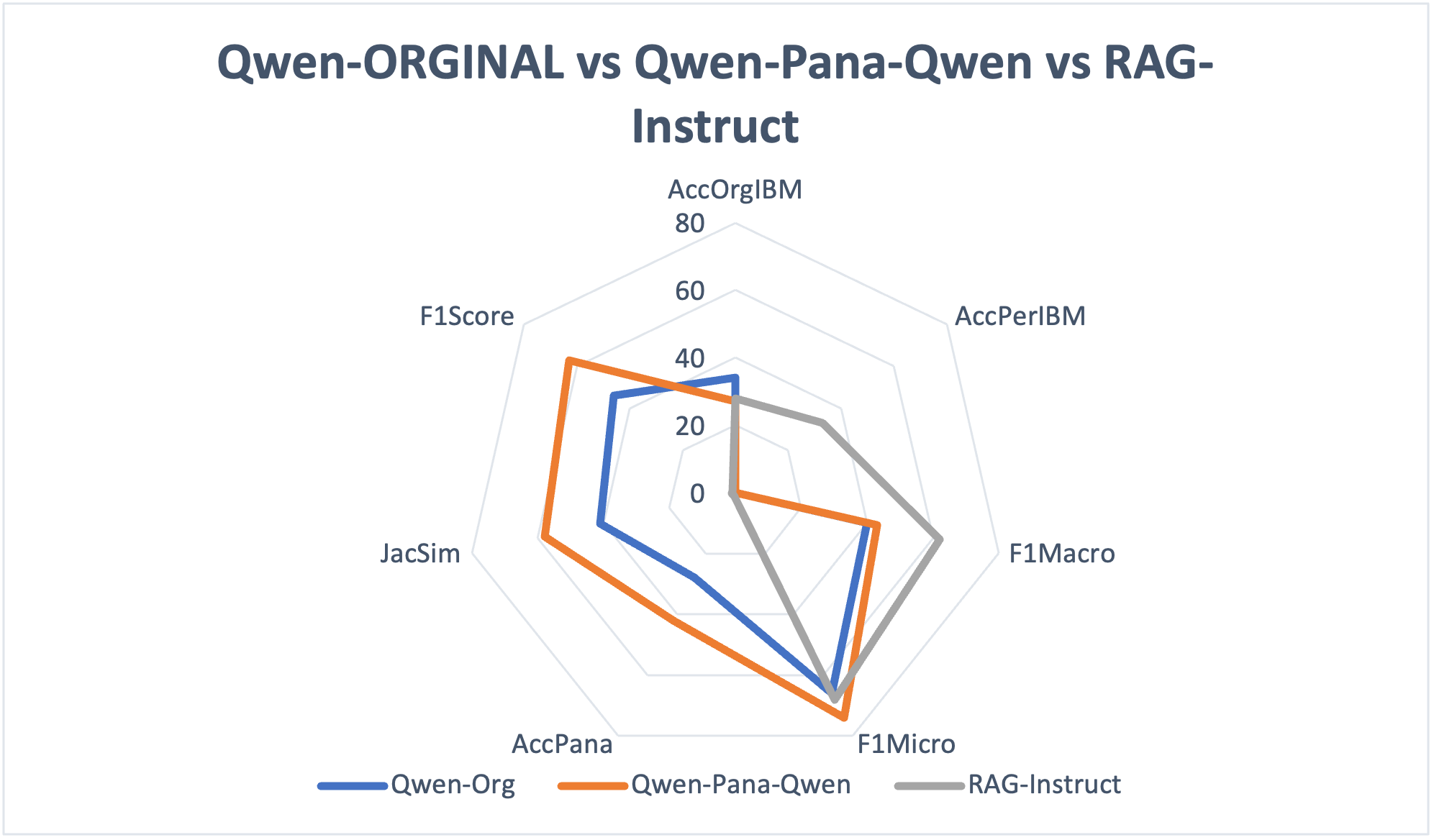}
    \caption{A side-by-side comparison of FailureSensorIQ and Panasonic metrics for Qwen3-4B-Instruct, Qwen3-4B-Instruct-Panasonic-Qwen, and RAG-Instruct-Llama3-8B.}
    \label{fig:comparison_side}
\end{figure}

Figure~\ref{fig:comparison_side} compares Qwen3-4B-Instruct (base), Qwen3-4B-Instruct-Panasonic-Qwen (fine-tuned on the Qwen3-30B-A3B-Instruct-generated dataset), and RAG-Instruct-Llama3-8B across four FailureSensorIQ metrics and three Panasonic metrics.

\begin{table}[t]
\small
\caption{Underlying values for Figure~\ref{fig:comparison_side}.}
\label{tab:comparison_qwen}
\centering
\scriptsize
\begin{tabular}{lccc}
\toprule
\textbf{Metric} & \textbf{Qwen-Org} & \textbf{Qwen-Pana-Qwen} & \textbf{RAG-Instruct} \\
\midrule
AccOrgIBM & 34\% & 27\% & 28\% \\
AccPerIBM & 0\% & 0\% & 33\% \\
F1-Macro (IBM) & 40\% & 43\% & 62\% \\
F1-Micro (IBM) & 66\% & 74\% & 68\% \\
Set-Match Acc. (Panasonic) & 28\% & 42\% & 1\% \\
Jaccard Similarity (Panasonic) & 41\% & 58\% & 1\% \\
F1-Score (Panasonic) & 46\% & 63\% & 1\% \\
\bottomrule
\end{tabular}
\end{table}

As with the Claude-generated dataset, fine-tuning on the Qwen3-30B-A3B-Instruct-generated dataset substantially improves all three Panasonic metrics (Set-Match Accuracy 28\%$\rightarrow$42\%, consistent with Table~\ref{tab:finetune_rag_results}), while RAG-Instruct-Llama3-8B remains essentially unusable on Panasonic (1\% across all three metrics), reproducing the severe drop already reported in Table~\ref{tab:base_model_results}. On FailureSensorIQ, however, the Qwen-generated dataset produces the opposite trade-off from the Claude-generated one: fine-tuning on Qwen data \emph{reduces} accuracy on the original, unperturbed questions (AccOrgIBM, 34\%$\rightarrow$27\%) while \emph{improving} both macro- and micro-averaged F1 (F1-Macro 40\%$\rightarrow$43\%; F1-Micro 66\%$\rightarrow$74\%) — a mirror image of the Claude-fine-tuned model, which gained AccOrgIBM but lost F1-Macro/F1-Micro (Section~\ref{sec:comparison_claude}). As with the Claude comparison, neither Qwen variant answers any perturbed FailureSensorIQ question correctly (AccPerIBM = 0\%), while RAG-Instruct-Llama3-8B is the strongest model on that metric (33\%) as well as on F1-Macro. This contrast between the two fine-tunes' effect on FailureSensorIQ is itself a notable finding: the choice of generator model does not just change the \emph{magnitude} of downstream gains, but can change \emph{which} FailureSensorIQ metrics improve versus regress.

\subsubsection{Claude-Generated Dataset}
\label{sec:comparison_claude}
\begin{figure}
    \centering
    \includegraphics[width=1.0\linewidth]{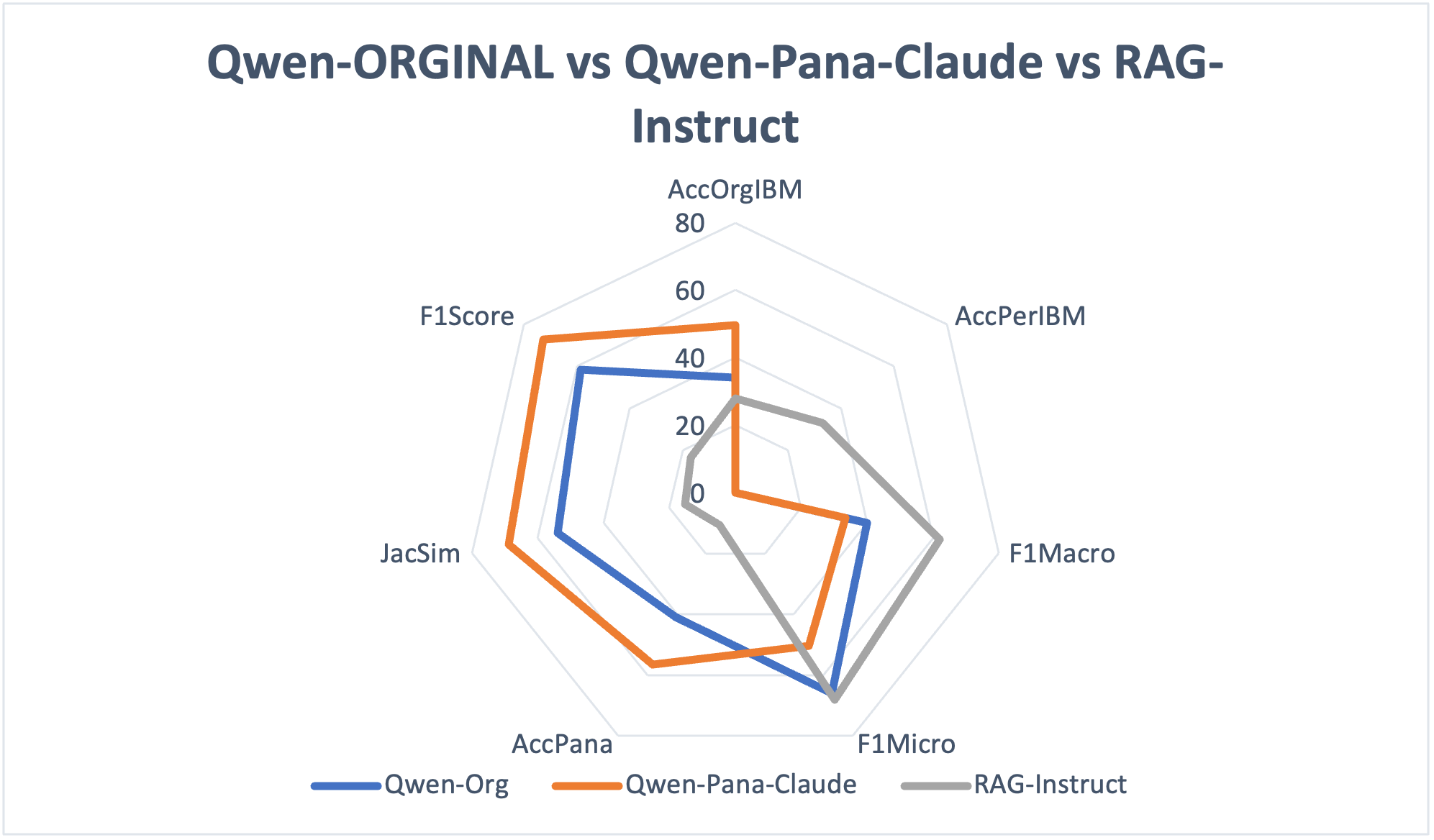}
    \caption{A side-by-side comparison of FailureSensorIQ and Panasonic metrics for Qwen3-4B-Instruct, Qwen3-4B-Instruct-Panasonic-Claude, and RAG-Instruct-Llama3-8B.}
    \label{fig:comparison_claude}
\end{figure}

Figure~\ref{fig:comparison_claude} compares Qwen3-4B-Instruct (base), Qwen3-4B-Instruct-Panasonic-Claude (fine-tuned on the Claude-Opus-4.6-generated dataset), and RAG-Instruct-Llama3-8B across four FailureSensorIQ metrics and three Panasonic metrics.

\begin{table}[t]
\small
\caption{Underlying values for Figure~\ref{fig:comparison_claude}.}
\label{tab:comparison_claude}
\centering
\scriptsize
\begin{tabular}{lccc}
\toprule
\textbf{Metric} & \textbf{Qwen-Org} & \textbf{Qwen-Pana-Claude} & \textbf{RAG-Instruct} \\
\midrule
AccOrgIBM & 34.0\% & 49.6\% & 28.0\% \\
AccPerIBM & 0.0\% & 0.0\% & 33.0\% \\
F1-Macro (IBM) & 40.0\% & 33.5\% & 62.0\% \\
F1-Micro (IBM) & 66.0\% & 50.3\% & 68.0\% \\
Set-Match Acc. (Panasonic) & 40.9\% & 56.4\% & 10.7\% \\
Jaccard Similarity (Panasonic) & 54.0\% & 68.9\% & 15.24\% \\
F1-Score (Panasonic) & 58.55\% & 72.66\% & 16.88\% \\
\bottomrule
\end{tabular}
\end{table}

Fine-tuning on the Claude-Opus-4.6-generated dataset improves Panasonic performance across all three metrics by a wide margin (e.g., Set-Match Accuracy 40.9\%$\rightarrow$56.4\%), consistent with Table~\ref{tab:claude_finetune_results}. On FailureSensorIQ, the picture is mixed: fine-tuning raises accuracy on the original, unperturbed questions (AccOrgIBM, 34.0\%$\rightarrow$49.6\%), but reduces both macro- and micro-averaged F1 (F1-Macro 40.0\%$\rightarrow$33.5\%; F1-Micro 66.0\%$\rightarrow$50.3\%), and neither the base nor the fine-tuned Qwen model answers any perturbed FailureSensorIQ question correctly (AccPerIBM = 0\% for both). RAG-Instruct-Llama3-8B shows the opposite profile: it is the strongest model on three of the four FailureSensorIQ metrics (AccPerIBM, F1-Macro, F1-Micro) yet collapses on Panasonic, scoring roughly 3--5$\times$ lower than either Qwen variant on all three Panasonic metrics (10.7\% Set-Match Accuracy vs. 40.9--56.4\%). This reinforces the pattern observed throughout this study: strength on one benchmark does not transfer to the other, and a model's retrieval-augmented performance is highly dependent on whether its underlying knowledge base and fine-tuning data match the target domain.

\subsubsection{Robustness to Perturbation: A Shared Limitation}
Across both comparisons, one result stands out for what fine-tuning could \emph{not} fix: AccPerIBM — accuracy on the perturbed, rephrased variant of FailureSensorIQ's questions — was 0\% for the base Qwen3-4B-Instruct model and remained 0\% after fine-tuning on \emph{either} the Qwen- or Claude-generated dataset (Tables~\ref{tab:comparison_qwen} and~\ref{tab:comparison_claude}). RAG-Instruct-Llama3-8B, by contrast, scored 33\% on this metric in both comparisons, despite performing far worse than either Qwen variant on the Panasonic benchmark. This suggests the failure is not simply a matter of domain knowledge — both fine-tuned models gained substantial Panasonic-specific knowledge — but reflects a more fundamental brittleness to question rephrasing in the Qwen3-4B backbone itself, one that neither of our generator models' training data was designed to address. This mirrors, and appears to sharpen, a pattern already reported for larger models on FailureSensorIQ, where perturbation was shown to reduce accuracy by as much as 41 percentage points~\cite{constantinides2025failuresensoriqmultichoiceqadataset}; here, for a small 4B-parameter model, the effect is total rather than partial. We view this as a limitation of the present study rather than a conclusion about fine-tuning data quality, and return to it in Section~\ref{sec:conclusion}.

\section{Conclusion and Future Directions}
\label{sec:conclusion}
This research provides a framework for constructing an appropriate dataset of industrial documents for model training and use in RAG. During this study, Panasonic product documents were collected and information and tables were extracted via an automated pipeline. We then artificially generated a dataset from these documents and evaluated its impact on the model. In this study, we used open-source models such as Dots.OCR and Qwen-30B-Instruct-2507, as well as the closed, API-based Claude-Opus-4.6, for dataset extraction and generation, and for evaluating impact, we employed Qwen-4B-Instruct, Phi-3-mini-4k, and the Llama3-based RAG-Instruct-8B. Also, in all of these studies, due to industry requirements for systems that retrieve all states, the simplest RAG method was implemented. The models used for the RAG system include Gemma3-300m for the embedding model and FAISS for retrieval. The results showed that this dataset can be effectively trained on small models with fewer than 10 billion parameters and that th s can improve model performance on evaluation metrics. Additionally, for dataset generation, it is possible to use open-source models, eliminating the need for massive commercial models like ChatGPT or Gemini. 

Generating the dataset with the open-weight Qwen3-30B-A3B-Instruct model on local hardware cost approximately \$3.2 in compute, compared with roughly \$330 in API usage for Claude-Opus-4.6 (Table~\ref{tab:cost_comparison}) — roughly two orders of magnitude more expensive. While Claude-Opus-4.6 produced a cleaner raw dataset (a far lower filtering rate, Section~\ref{sec:claude_regeneration}) and yielded larger downstream fine-tuning gains (Section~\ref{sec:claude_training}), the improvement was not proportional to the cost difference, suggesting that open-weight models remain a highly cost-effective choice for this type of dataset construction. Moreover, the open-weight model ecosystem is advancing rapidly, and larger open-weight models are increasingly capable of approaching the quality of closed, API-based frontier models at a fraction of the cost.

\begin{table}[t]
\scriptsize
\caption{Compute/monetary cost of generating the Panasonic dataset with each generator model.}
\label{tab:cost_comparison}
\centering
\begin{tabular}{@{}p{2.3cm}p{1.3cm}p{1.0cm}p{2.4cm}@{}}
\toprule
\textbf{Generator} & \textbf{Compute Time} & \textbf{Cost} & \textbf{Notes} \\
\midrule
Qwen3-30B-A3B-Instruct & 1h 43m & \$3.2 & Local Pro 6000 WS \\
Claude-Opus-4.6 & --- & \$330 & Token-based billing \\
\bottomrule
\end{tabular}
\end{table}

For future work, we plan to evaluate a broader range of language models to further analyze generalization across architectures and parameter scales. In addition, we aim to expand the dataset by incorporating documents from diverse industrial knowledge sources in order to construct a large-scale, comprehensive industrial benchmark. Future research will also explore more advanced RAG architectures beyond the basic implementation used in this study. We also intend to investigate multimodal industrial documents that integrate textual and visual information, enabling the extraction and utilization of embedded images, diagrams, and structured visual content alongside text to improve model robustness in real-world industrial scenarios. Most notably, every model evaluated in this study — regardless of fine-tuning data — completely failed to answer any perturbed FailureSensorIQ question correctly (Section~\ref{sec:industrial_comparison}), a limitation our current dataset construction pipeline does not address since it does not include rephrased or perturbed variants of its own generated questions. We consider closing this gap, for instance by explicitly generating paraphrased or adversarially rephrased variants of each Q\&A pair during dataset construction, one of the most promising directions for improving robustness in future iterations of Industrial-Instruction.

\appendix

\section{Dataset Creation Process}

The dataset construction process was completed in a total of 1 hour and 43 minutes on a Pro 6000 WS system.

\section{Model Training Settings and Time}

The model was trained for 12 hours, 3 minutes, and 3 seconds using two NVIDIA RTX 5090 GPUs. The target batch size during training was set to 64. Due to GPU memory limitations, directly implementing this batch size was not feasible; therefore, gradient accumulation was employed. In this configuration, the per-device batch size was set to 2 with 32 gradient accumulation steps, resulting in an effective batch size of 64. It should be noted that, in this setup, the computed loss represents an accumulated estimate and may slightly differ from the exact large-batch gradient.

\section{Additional MMLU Detail}
\label{app:mmlu_detail}
Table~\ref{tab:mmlu_detail_qwen} and Table~\ref{tab:mmlu_detail_claude} report the largest subject-level MMLU changes for finetuned\_v1\_qwen and finetuned\_v2\_claude, respectively, relative to the base model. Full 57-subject scores for all three models are available in our released evaluation logs.

\begin{table}[t]
\small
\caption{Largest MMLU regressions for finetuned\_v1\_qwen (subjects with the biggest accuracy drop vs. the base model).}
\label{tab:mmlu_detail_qwen}
\centering
\begin{tabular}{lccc}
\toprule
\textbf{Subject} & \textbf{Original} & \textbf{finetuned\_v1\_qwen} & \textbf{$\Delta$} \\
\midrule
moral\_scenarios & 51.17\% & 40.45\% & \textbf{-10.72} \\
public\_relations & 69.09\% & 62.73\% & -6.36 \\
marketing & 91.88\% & 87.18\% & -4.70 \\
high\_school\_biology & 90.65\% & 86.13\% & -4.52 \\
high\_school\_microeconomics & 91.60\% & 87.82\% & -3.78 \\
human\_aging & 73.54\% & 69.96\% & -3.58 \\
anatomy & 71.11\% & 68.15\% & -2.96 \\
astronomy & 87.50\% & 84.87\% & -2.63 \\
\bottomrule
\end{tabular}
\end{table}

\begin{table}[t]
\small
\caption{Largest MMLU changes for finetuned\_v2\_claude vs. the base model: regressions (top) and improvements (bottom).}
\label{tab:mmlu_detail_claude}
\centering
\begin{tabular}{lccc}
\toprule
\textbf{Subject} & \textbf{Original} & \textbf{finetuned\_v2\_claude} & \textbf{$\Delta$} \\
\midrule
public\_relations & 69.09\% & 63.64\% & -5.45 \\
anatomy & 71.11\% & 66.67\% & -4.44 \\
college\_chemistry & 54.00\% & 50.00\% & -4.00 \\
high\_school\_computer\_science & 89.00\% & 85.00\% & -4.00 \\
marketing & 91.88\% & 88.46\% & -3.42 \\
\midrule
global\_facts & 43.00\% & 51.00\% & \textbf{+8.00} \\
college\_mathematics & 48.00\% & 55.00\% & +7.00 \\
machine\_learning & 60.71\% & 66.07\% & +5.36 \\
security\_studies & 74.29\% & 78.78\% & +4.49 \\
international\_law & 80.17\% & 83.47\% & +3.30 \\
\bottomrule
\end{tabular}
\end{table}

Notably, \texttt{public\_relations} and \texttt{anatomy} regress under \emph{both} fine-tunes, which suggests these two subjects may be inherently sensitive to fine-tuning on this base model regardless of the training data source, rather than an artifact specific to either dataset. Conversely, a handful of subjects — \texttt{college\_mathematics}, \texttt{machine\_learning}, and \texttt{moral\_disputes} among them — improve under \emph{both} fine-tunes, suggesting the fine-tuning process itself contributes some dataset-independent generalization alongside whatever forgetting occurs elsewhere. We recommend manually inspecting the \texttt{moral\_scenarios} log samples for \texttt{finetuned\_v1\_qwen} specifically, since a -10.7 point swing on a single subject is large enough to warrant ruling out answer-format drift as opposed to genuine reasoning loss.

\section{Prompt Templates for the Five RAG Scenarios}
\label{app:prompts}

Each of the five dataset-generation scenarios (r0--r4, Section~\ref{sec:dataset_creation}) uses a prompt built from a shared template that instructs Qwen3-30B-A3B-Instruct to generate a question $q^{*}$, an answer $a^{*}$, and, when the simulated instruction is multiple-choice, five options labeled A--E. All five prompts share the same document placeholder(s), instruction-simulation mechanism, and output format; they differ only in how many documents are supplied and in the single requirement rule that defines the query--document relationship, as summarized in Table~\ref{tab:prompt_diff}.

\begin{table}[H]
\small
\caption{How each scenario prompt differs from the shared template.}
\label{tab:prompt_diff}
\centering
\begin{tabular}{lp{1.3cm}p{4.3cm}}
\toprule
\textbf{Scenario} & \textbf{Docs} & \textbf{Core requirement (rule 1)} \\
\midrule
Useless Document (r0) & 1 & Document is related but provides no useful information for answering $q^{*}$. \\
Single-Doc.\ Support (r1) & 1 & Document provides supporting clues but not an explicit answer. \\
Multi-Doc.\ Support (r2) & $\geq$2 & Multiple documents jointly provide clues but not an explicit answer. \\
Single-Doc.\ Answer (r3) & 1 & Document contains the complete, explicit answer. \\
Multi-Doc.\ Answer (r4) & $\geq$2 & Answer requires multi-hop reasoning across documents. \\
\bottomrule
\end{tabular}
\end{table}

Listing~\ref{lst:prompt_r0} shows the full prompt used for the Useless Document (r0) scenario as a representative example; the remaining four prompts follow the identical structure with only the document count and rule 1 changed as listed in Table~\ref{tab:prompt_diff}.

\begin{lstlisting}[style=promptstyle, float=tbp, caption={Full prompt template for scenario r0 (Useless Document).}, label={lst:prompt_r0}]
<Documents>
{doc}
</Documents>
Your task is to generate an English question q* and a corresponding response a* based on the provided <Documents>. Please note that the question q* can take various forms, not limited to questions with a question mark, but also including statements, instructions, and other formats. You need to follow the requirements below to generate the q* and a* (RAG Paradigms):
1. q* should be related to the <Documents>, but the <Documents> can not provide any useful information for answering q*.
2. a* should be able to answer q*, ensuring that the response a* is accurate, detailed, and comprehensive.
3. q* must be a standalone question. Strictly avoid using phrases that reference the source material, such as "Based on the provided context," "According to the documents," "In the text," or similar meta-references.
Additionally, to ensure diversity, richness, and high quality in the question q* you generate, we will randomly provide a question for you to emulate. In other words, while satisfying the requirements above, make q* similar in task requirement and expression to the <Simulated Instruction> below:
<Simulated Instruction>
{simulated_instruction}
</Simulated Instruction>
IMPORTANT: If the <Simulated Instruction> is a multiple-choice question with options, then your generated q* MUST ALSO include exactly 5 options labeled A-E, and a* MUST include the correct option(s) in the same JSON list format used in the simulated instruction. Always generate 5 options, even if the original document does not contain any option-like content.
Please directly generate the question-answer-options (q*, a*, options*) following all the rules above in the format of {"q*": ..., "a*": ..., "options*": ...}, so for a* select A-E correct option.
Ensure the quality of the generated (q*, a*, options*).
\end{lstlisting}

Full source code for all five prompt templates, along with the dataset generation pipeline, is available in our public code repository.

\section*{Acknowledgments}
We thank Hamedan University of Technology and the University of Antwerp / Flanders Make Strategic Research Center for their institutional support of this work.

\bibliographystyle{plainnat}
\bibliography{References}

\end{document}